# Learning the Past Tense of English Verbs: The Symbolic Pattern Associator vs. Connectionist Models

**Charles X. Ling** LING@CSD.UWO.CA
*Department of Computer Science*
*The University of Western Ontario*
*London, Ontario, Canada N6A 5B7*

## Abstract

Learning the past tense of English verbs — a seemingly minor aspect of language acquisition — has generated heated debates since 1986, and has become a landmark task for testing the adequacy of cognitive modeling. Several artificial neural networks (ANNs) have been implemented, and a challenge for better symbolic models has been posed. In this paper, we present a general-purpose Symbolic Pattern Associator (SPA) based upon the decision-tree learning algorithm ID3. We conduct extensive *head-to-head* comparisons on the generalization ability between ANN models and the SPA under different representations. We conclude that the SPA generalizes the past tense of unseen verbs better than ANN models by a wide margin, and we offer insights as to why this should be the case. We also discuss a new default strategy for decision-tree learning algorithms.

## 1. Introduction

Learning the past tense of English verbs, a seemingly minor aspect of language acquisition, has generated heated debates since the first connectionist implementation in 1986 (Rumelhart & McClelland, 1986). Based on their results, Rumelhart and McClelland claimed that the use and acquisition of human knowledge of language can best be formulated by ANN (Artificial Neural Network) models without symbol processing that postulates the existence of explicit symbolic representation and rules. Since then, learning the past tense has become a landmark task for testing the adequacy of cognitive modeling. Over the years a number of criticisms of connectionist modeling appeared (Pinker & Prince, 1988; Lachter & Bever, 1988; Prasada & Pinker, 1993; Ling, Cherwenka, & Marinov, 1993). These criticisms centered mainly upon the issues of high error rates and low reliability of the experimental results, the inappropriateness of the training and testing procedures, "hidden" features of the representation and the network architecture that facilitate learning, as well as the opaque knowledge representation of the networks. Several subsequent attempts at improving the original results with new ANN models have been made (Plunkett & Marchman, 1991; Cottrell & Plunkett, 1991; MacWhinney & Leinbach, 1991; Daugherty & Seidenberg, 1993). Most notably, MacWhinney and Leinbach (1991) constructed a multilayer neural network with backpropagation (BP), and attempted to answer early criticisms. On the other hand, supporters of the symbolic approach believe that symbol structures such as parse trees, propositions, etc., and the rules for their manipulations, are critical at the cognitive level, while the connectionist approach may only provide an account of the neural structures in which the traditional symbol-processing cognitive architecture is implemented (Fodor & Pylyshyn, 1988). Pinker (1991) and Prasada and Pinker (1993) argue that a proper accounting for regular verbs should be dependent upon production rules, while irregular past-tense inflections may be generalized by ANN-like associative memory.

The proper way of debating the adequacy of symbolic and connectionist modeling is by contrasting competitive implementations. Thus, a symbolic implementation is needed that can be compared with the ANN models. This is, in fact, a challenge posed by MacWhinney and Leinbach (1991), who assert that no symbolic methods would work as well as their own model. In a section titled "Is there a better symbolic model?" they claim:

> If there were some other approach that provided an even more accurate characterization of the learning process, we might still be forced to reject the connectionist approach, despite its successes. The proper way of debating conceptualizations is by contrasting competitive implementations. To do this in the present case, we would need a symbolic implementation that could be contrasted with the current implementation. (MacWhinney & Leinbach, 1991, page 153)

In this paper, we present a general-purpose Symbolic Pattern Associator (SPA) based upon the symbolic decision tree learning algorithm ID3 (Quinlan, 1986). We have shown (Ling & Marinov, 1993) that the SPA's results are much more psychologically realistic than ANN models when compared with human subjects. On the issue of the predictive accuracy, MacWhinney and Leinbach (1991) did not report important results of their model on unseen regular verbs. To reply to our criticism, MacWhinney (1993) re-implemented the ANN model, and claimed that its raw generalization power is very close to that of our SPA. He believed that this should be the case because both systems learn from the same data set:

> There is a very good reason for the equivalent performance of these two models. [...] When two computationally powerful systems are given the same set of input data, they both extract every bit of data regularity from that input. Without any further processing, there is only so much blood that can be squeezed out of a turnip, and each of our systems [SPA and ANN] extracted what they could. (MacWhinney, 1993, page 295)

We will show that this is not the case; obviously there *are* reasons why one learning algorithm outperforms another (otherwise why do we study different learning algorithms?). The Occam's Razor Principle — preferring the simplest hypothesis over more complex ones — creates preference biases for learning algorithms. A *preference bias* is a preference order among competitive hypotheses in the hypothesis space. Different learning algorithms, however, employ different ways of measuring simplicity, and thus concepts that they bias to are different. How well a learning program generalizes depends upon the degree to which the regularity of the data fits with its bias. We study and compare the raw generalization ability of symbolic and ANN models on the task of learning the past tense of English verbs. We perform extensive head-to-head comparisons between ANN and SPA, and show the effects of different representations and encodings on their generalization abilities. Our experimental results demonstrate clearly that

1. the distributed representation, a feature that connectionists have been advocating, does not lead to better generalization when compared with the symbolic representation, or with arbitrary error-correcting codes of a proper length;

2. ANNs cannot learn the identity mapping that preserves the verb stem in the past tense as well as the SPA can;

3. a new representation suggested by MacWhinney (1993) improves the predictive accuracy of both SPA and ANN, but SPA still outperforms ANN models;

4. in sum, the SPA generalizes the past tense of unseen verbs better than ANN models by a wide margin.

In Section 5 we discuss reasons as to why the SPA is a better learning model for the the task of English past-tense acquisition. Our results support the view that many such rule-governed cognitive processes should be better modeled by symbolic, rather than connectionist, systems.

## 2. Review of Previous Work

In this section, we review briefly the two main connectionist models of learning the past tenses of English verbs, and the subsequent criticisms.

### 2.1 Rumelhart and McClelland's Model

Rumelhart and McClelland's model is based on a simple perceptron-based pattern associator interfaced with an input/output encoding/decoding network which allows the model to associate verb stems with their past tenses using a special Wickelphone/Wickelfeature phoneme-representation format. The learning algorithm is the classical perceptron convergence procedure. The training and the testing sets are mutually disjoint in the experiments. The errors made by the model during the training process broadly follow the U-shaped learning curve in the stages of acquisition of the English past tense exhibited by young children. The testing sample consists of 86 "unseen" low frequency verbs (14 irregular and 72 regular) that are not randomly chosen. The testing sample results have a 93% error rate for the irregulars. The regulars fare better with a 33.3% error rate. Thus, the overall error rate for the whole testing sample is 43% — 37 wrong or ambiguous past tense forms out of 86 tested. Rumelhart and McClelland (1986) claim that the outcome of their experiment disconfirms the view that there exist explicit (though inaccessible) rules that underlie human knowledge of language.

### 2.2 MacWhinney and Leinbach's Model

MacWhinney and Leinbach (1991) report a new connectionist model on the learning of the past tenses of English verbs. They claim that the results from the new simulation are far superior to Rumelhart and McClelland's results, and that they can answer most of the criticisms aimed at the earlier model. The major departure from Rumelhart and McClelland's model is that the Wickelphone/Wickelfeature representational format is replaced with the UNIBET (MacWhinney, 1990) phoneme representational system which allows the assignment of a *single* alphabetic/numerical letter to each of the total 36 phonemes. MacWhinney and Leinbach use special *templates* with which to code each phoneme and its position in a word. The actual input to the network is created by coding the individual phonemes as sets of phonetic features in a way similar to the coding of Wickelphones as Wickelfeatures (cf Section 4.3). The network has two layers of 200 "hidden" units fully connected to adjacent layers. This number was arrived at through trial and error. In addition, the network has a special-purpose set of connections that copy the input units directly onto the output units.

Altogether, 2062 regular and irregular English verbs are selected for the experiment — 1650 of them are used for training (1532 regular and 118 irregular), but only 13 low frequency irregular verbs are used for testing (MacWhinney & Leinbach, 1991, page 144). Training the network takes 24,000 epochs. At the end of training there still are 11 errors on the irregular pasts. MacWhinney and Leinbach believe that if they allow the network to run for several additional days and give it additional hidden unit resources, it probably can reach complete convergence (MacWhinney & Leinbach, 1991, page 151). The only testing error rate reported is based on a very small and biased test sample of 13 unseen irregular verbs; 9 out of 13 are predicted incorrectly. They do not test their model on any of the unseen regular verbs: "Unfortunately, we did not test a similar set of 13 regulars." (MacWhinney & Leinbach, 1991, page 151).

### 2.3 Criticism of the Connectionist Models

Previous and current criticisms of the connectionist models of learning the past tenses of English verbs center mainly on several issues. Each of these issues is summarized in the following subsections.

#### 2.3.1 Error Rates

The error rate in producing the past tenses of the "unseen" test verbs is very high in both ANN models, and important tests were not carried out in MacWhinney and Leinbach (1991) model. The experimental results indicate that neither model reaches the level of *adult competence*. In addition, relatively large numbers of the errors are not psychologically realistic since humans rarely make them.

#### 2.3.2 Training and Testing Procedures

In both Rumelhart and McClelland's model, and MacWhinney and Leinbach's model, the generalization ability is measured on only *one* training/testing sample. Further, the testing sets are not randomly chosen, and they are very small. The accuracy in testing irregular verbs can vary greatly depending upon the particular set of testing verbs chosen, and thus multiple runs with large testing samples are necessary to assess the true generalization ability of a learning model. Therefore, the results of the previous connectionist models are not reliable. In Section 4, we set up a reliable testing procedure to compare connectionist models with our symbolic approach. Previous connectionist simulations have also been criticized for their crude training processes (for example, the sudden increase of regular verbs in the training set), which create such behavior as the U-shaped learning curves.

#### 2.3.3 Data Representation and Network Architecture

Most of the past criticisms of the connectionist models have been aimed at the data-representation formats employed in the simulations. Lachter and Bever (1988) pointed out that the results achieved by Rumelhart and McClelland's model would have been impossible without the use of several TRICS (The Representations It Crucially Supposes) introduced with the adoption of the Wickelphone/Wickelfeature representational format. MacWhinney and Leinbach claim that they have improved upon the earlier connectionist model by getting rid of the Wickelphone/Wickelfeature representation format, and thus to have responded to the many criticisms that this format entailed. However, MacWhinney and Leinbach also introduce several TRICS in their data-representation format. For example, instead of coding predecessor and successor phonemes as Wickelphones, they introduce special templates with which to code positional information. This means that the network will learn to associate patterns of phoneme/positions within a predetermined consonant/vowel pattern. Further, the use of restrictive templates gets rid of many English verbs that do not fit the chosen template. This may bias the model in favour of shorter verbs, predominantly of Anglo-Saxon origin, and against longer verbs, predominantly composite or of Latin and French origin. Another TRICS introduced is the phonetic feature encoding (a distributed representation). It is not clear why phonetic features such as front, centre, back, high, etc. are chosen. Do they represent finer grained "microfeatures" that help to capture the regularities in English past tenses? In Section 4.5, we will show that the straightforward symbolic representation leads to better generalization than does the carefully engineered distributed representation. This undermines the claimed advantages of the distributed representation of connectionist models.

#### 2.3.4 Knowledge Representation and Integration of Acquired Knowledge

Pinker and Prince (1988), and Lachter and Bever (1988) point out that Rumelhart and McClelland try to model the acquisition of the production of the past tense in *isolation* from the rest of the English morphological system. Rumelhart and McClelland, as well as MacWhinney and Leinbach, assume that the acquisition process establishes a direct mapping from the phonetic representation of the stem to the phonetic representation of the past tense form. This direct mapping collapses some well-established distinctions such as lexical item vs. phoneme string, and morphological category vs. morpheme. Simply remaining at the level of phonetic patterns, it is impossible to express new categorical information in first-order (predicate/function/variable) format. One of the inherent deficits of the connectionist implementations is that there is no such thing as a variable for *verb stem*, and hence there is no way for the model to attain the knowledge that one could add suffix to a *stem* to get its past tense (Pinker & Prince, 1988, page 124). Since the acquired knowledge in such networks is a large weight matrix, which usually is opaque to the human observer, it is unclear how the phonological levels processing that the connectionist models carry out can be integrated with the morphological, lexical, and syntactical level of processing. Neither Rumelhart and McClelland nor MacWhinney and Leinbach address this issue. In contrast to ANNs whose internal representations are entirely opaque, the SPA can represent the acquired knowledge in the form of production rules, and allow for further processing, resulting in higher-level categories such as the verb stem and the voiced consonants, linguistically realistic production rules using these new categories for regular verbs, and associative templates for irregular verbs (Ling & Marinov, 1993).

## 3. The Symbolic Pattern Associator

We take up MacWhinney and Leinbach's challenge for a better symbolic model for learning the past tense of English verbs, and present a general-purpose Symbolic Pattern Associator (SPA)[1] that can generalize the past tense of unseen verbs much more accurately than connectionist models in this section. Our model is *symbolic* for several reasons. First, the input/output representation of the learning program is a set of phoneme symbols, which are the basic elements governing the past-tense inflection. Second, the learning program operates on those phoneme symbols directly, and the acquired knowledge can be represented in the form of production rules using those phoneme symbols as well. Third, those production rules at the phonological level can easily be further generalized into first-order rules that use more abstract, high-level symbolic categories such as morphemes and the verb stem (Ling & Marinov, 1993). In contrast, the connectionist models operate on a distributed representation (phonetic feature vectors), and the acquired knowledge is embedded in a large weight matrix; it is therefore hard to see how this knowledge can be further generalized into more abstract representations and categories.

### 3.1 The Architecture of the Symbolic Pattern Associator

The SPA is based on C4.5 (Quinlan, 1993) which is an improved implementation of the ID3 learning algorithm (cf. (Quinlan, 1986)). ID3 is a program for inducing classification rules in the form of decision trees from a set of classified examples. It uses information gain ratio as a criterion for selecting attributes as roots of the subtrees. The divide-and-conquer strategy is recursively applied in building subtrees until all remaining examples in the training set belong to a single concept (class); then a leaf is labeled as that concept. The information gain guides a greedy heuristic search for the locally *most relevant* or *discriminating* attribute that maximally reduces the entropy (randomness) in the divided set of the examples. The use of this heuristic usually results in building *small* decision trees instead of larger ones that also fit the training data.

If the task is to learn to classify a set of different patterns into a single class of several mutually exclusive categories, ID3 has been shown to be comparable with neural networks (i.e., within about 5% range on the predictive accuracy) on many real-world learning tasks (cf. (Shavlik, Mooney, & Towell, 1991; Feng, King, Sutherland, & Henery, 1992; Ripley, 1992; Weiss & Kulikowski, 1991)). However, if the task is to classify a set of (input) patterns into (output) patterns of many attributes, ID3 cannot be applied directly. The reason is that if ID3 treats the different output patterns as mutually exclusive classes, the number of classes would be exponentially large and, more importantly, any generalization of individual output attributes within the output patterns would be lost.

To turn ID3 or any similar N-to-1 classification system into general purpose N-to-M symbolic pattern associators, the SPA applies ID3 on all output attributes and combines individual decision trees into a "forest", or set of trees. A similar approach was proposed for dealing with the distributed (binary) encoding in multiclass learning tasks such as NETtalk (English text-to-speech mapping) (Dietterich, Hild, & Bakiri, 1990). Each tree takes as input the set of all attributes in the input patterns, and is used to determine the value of

1. The SPA programs and relevant datasets can be obtained anonymously from ftp.csd.uwo.ca under pub/SPA/ .

one attribute in its output pattern. More specifically, if a pair of input attributes ($\iota_1$ to $\iota_n$) and output attributes ($\omega_1$ to $\omega_m$) is represented as:

$$\iota_1, ..., \iota_n \rightarrow \omega_1, ..., \omega_m$$

then the SPA will build a total of $m$ decision trees, one for each output attribute $\omega_i$ ($1 \leq i \leq m$) taking all input attributes $\iota_1, ..., \iota_n$ per tree. Once all of $m$ trees are built, the SPA can use them jointly to determine the output pattern $\omega_1, ..., \omega_m$ from any input pattern $\iota_1, ..., \iota_n$.

An important feature of the SPA is explicit knowledge representation. Decision trees for output attributes can easily be transformed into propositional production rules (Quinlan, 1993). Since entities of these rules are symbols with semantic meanings, the acquired knowledge often is comprehensible to the human observer. In addition, further processing and integration of these rules can yield high-level knowledge (e.g., rules using verb stems) (Ling & Marinov, 1993). Another feature of the SPA is that the trees for different output attributes contain identical components (branches and subtrees) (Ling & Marinov, 1993). These components have similar roles as hidden units in ANNs since they are shared in the decision trees of more than one output attribute. These identical components can also be viewed as high-level concepts or feature combinations created by the learning program.

### 3.2 Default Strategies

An interesting research issue is how decision-tree learning algorithms handle the default class. A *default class* is the class to be assigned to leaves which no training examples are classified into. We call these leaves *empty leaves*. This happens when the attributes have many different values, or when the training set is relatively small. In these cases, during the tree construction, only a few branches are explored for some attributes. When the testing examples fall into the empty leaves, a *default strategy* is needed to assign classes to those empty leaves.

For easier understanding, we use the spelling form of verbs in this subsection to explain how different default strategies work. (In the actual learning experiment the verbs are represented in phonetic form.) If we use consecutive left-to-right alphabetic representation, the verb stems and their past tenses of a small training set can be represented as follows:

```
a,f,f,o,r,d,_,_,_,_,_,_,_,_,_ => a,f,f,o,r,d,e,d,_,_,_,_,_,_,_
e,a,t,_,_,_,_,_,_,_,_,_,_,_,_ => a,t,e,_,_,_,_,_,_,_,_,_,_,_,_
l,a,u,n,c,h,_,_,_,_,_,_,_,_,_ => l,a,u,n,c,h,e,d,_,_,_,_,_,_,_
l,e,a,v,e,_,_,_,_,_,_,_,_,_,_ => l,e,f,t,_,_,_,_,_,_,_,_,_,_,_
```

where _ is used as a filler for empty space. The left-hand 15 columns are the input patterns for the stems of the verbs; the right-hand 15 columns are the output patterns for their corresponding correct past tense forms.

As we have discussed, 15 decision trees will be constructed, one for each output attribute. The decision tree for the first output attribute can be constructed (see Figure 1 (a)) from the following 4 examples:

```
a,f,f,o,r,d,_,_,_,_,_,_,_,_,_ => a
e,a,t,_,_,_,_,_,_,_,_,_,_,_,_ => a
l,a,u,n,c,h,_,_,_,_,_,_,_,_,_ => l
```

```
l,e,a,v,e,_,_,_,_,_,_,_,_,_,_ => l
```

where the last column is the classification of the first output attribute. However, many other branches (such as $\iota_1 = c$ in Figure 1 (a)) are not explored, since no training example has that attribute value. If a testing pattern has its first input attribute equal to $c$, what class should it be assigned to? ID3 uses the *majority default.* That is, the most popular class in the whole subtree under $\iota_1$ is assigned to the empty leaves. In the example above, either class *a* or *l* will be chosen since they each have 2 training examples. However, this is clearly not the right strategy for this task since a verb such as *create* would be output as *l......* or *a......*, which is incorrect. Because it is unlikely for a small training set to have all variations of attribute values, the majority default strategy of ID3 is not appropriate for this task.

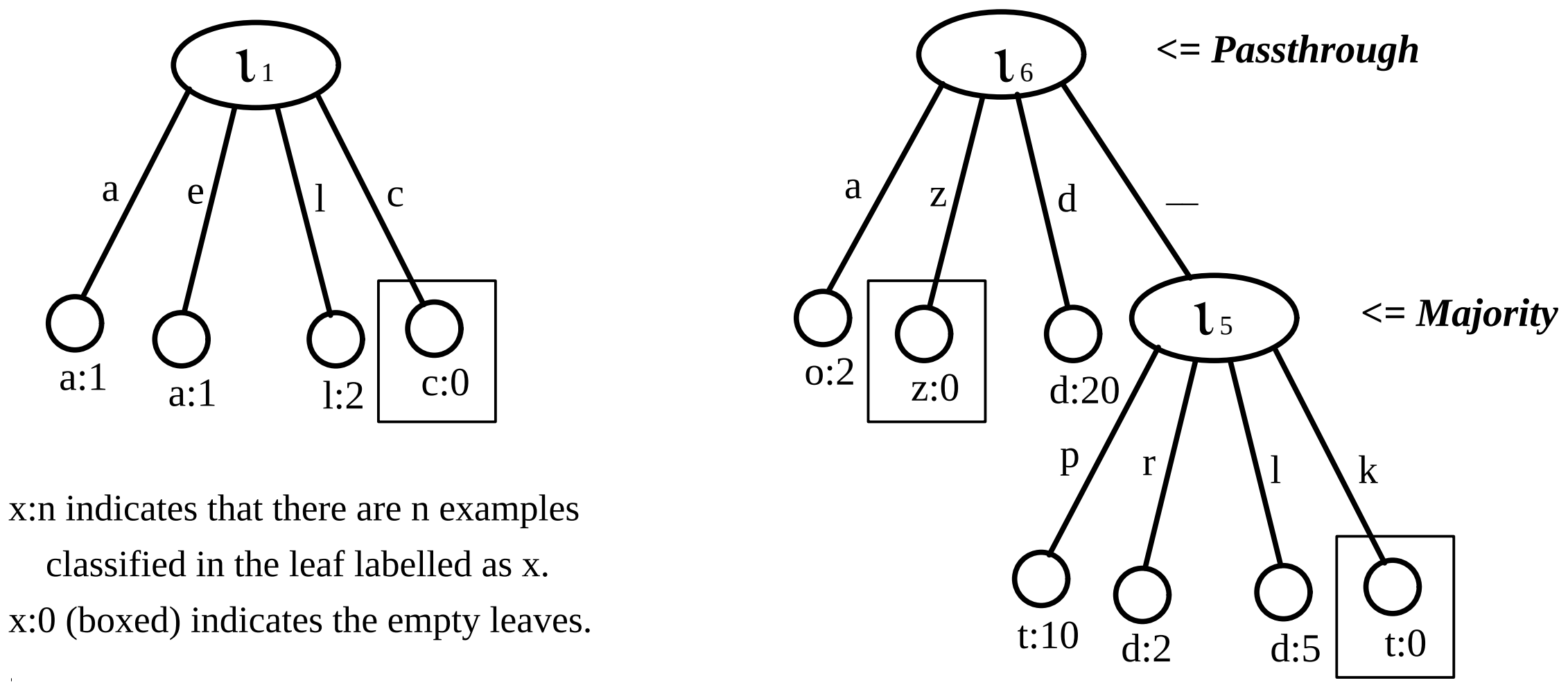

Figure 1: (a) Passthrough default (b) Various default

For applications such as verb past-tense learning, a new default heuristic — *passthrough* — may be more suitable. That is, the classification of an empty leaf should be the same as the attribute value of that branch. For example, using the passthrough default strategy, *create* will be output as *c.......* The passthrough strategy gives decision trees some first-order flavor, since the production rules for empty leaves can be represented as *If Attribute* $= X$ *then Class* $= X$ where $X$ can be any unused attribute values. Passthrough is a domain-dependent heuristic strategy because the class labels may have nothing to do with the attribute values in other applications.

Applying the passthrough strategy alone, however, is not adequate for every output attribute. The endings of the regular past tenses are not identical to any of the input patterns, and the irregular verbs may have vowel and consonant changes in the middle of the verbs. In these cases, the majority default may be more suitable than the passthrough. In order to choose the right default strategy — majority or passthrough — a decision is made based upon the training data in the corresponding subtree. The SPA first determines the majority class, and counts the number of examples from all subtrees that belong to this class. It then counts the number of examples in the subtrees that coincide with the passthrough strategy. These two numbers are compared, and the default strategy employed by more examples is chosen. For instance, in the example above (see Figure 1 (a)), the majority class is *l* (or a) having 2 instances. However, there are 3 examples coinciding with the passthrough default: two *l* and one *a*. Thus the passthrough strategy takes over, and assigns all empty leaves *at this level*. The empty attribute branch *c* would then be assigned the class *c*. Note that the default strategy for empty leaves of attribute $X$ depends upon training examples falling into the subtree rooted at $X$. This localized method ensures that only related objects have an influence on calculating default classes. As a result, the SPA can *adapt* the default strategy that is best suited at different levels of the decision trees. For example, in Figure 1 (b), two different default strategies are used at different levels in the same tree. We use the SPA with the adaptive default strategy throughout the remainder of this paper. Note that the new default strategy is not a TRICS in the data representation; rather, it represents a bias of the learning program. Any learning algorithm has a default strategy independent of the data representation. The effect of different data representations on generalization is discussed in Sections 4.3, 4.5, and 4.6. The passthrough strategy can be imposed on ANNs as well by adding a set of copy connections between the input units and the twin output units. See Section 4.4 for detail.

### 3.3 Comparisons of Default Strategies of ID3, SPA, and ANN

Which default strategy do neural networks tend to take in generalizing default classes when compared with ID3 and SPA? We conducted several experiments to determine neural networks' default strategy. We assume that the domain has only one attribute $X$ which may take values *a*, *b*, *c*, and *d*. The class also can be one of the *a*, *b*, *c*, and *d*. The training examples have attribute values *a*, *b*, and *c* but not *d* — it is reserved for testing the default class. The training set contains multiple copies of the same example to form a certain majority class. Table 1 shows two sets of training/testing examples that we used to test and compare default strategies of ID3, SPA and neural networks.

| Data set 1 | | | Data set 2 | | |
|---|---|---|---|---|---|
| Training examples | | | Training examples | | |
| Values of $X$ | Class | # of copies | Values of $X$ | Class | # of copies |
| a | a | 10 | a | c | 10 |
| b | b | 2 | b | b | 6 |
| c | c | 3 | c | c | 7 |
| Testing example | | | Testing example | | |
| d | ? | 1 | d | ? | 1 |

Table 1: Two data sets for testing default strategies of various methods.

The classification of the testing examples by ID3 and SPA is quite easy to decide. Since ID3 takes only the majority default, the output class is *a* (with 10 training examples) for the first data set, and *c* (with 17 training examples) for the second data set. For SPA, the number of examples using passthrough is 15 for the first data set, and 13 for the second data set. Therefore, the passthrough strategy wins in the first case with the output class $d$, and the majority strategy wins in the second case with the output class $c$.

For neural networks, various coding methods were used to represent values of the attribute $X$. In the dense coding, we used 00 to represent $a$, 01 for $b$, 10 for $c$ and 11 for $d$. We also tried the standard one-per-class encoding, and real number encoding (0.2 for $a$, 0.4 for $b$, 0.6 for $c$ and 0.8 for $d$). The networks were trained using as few hidden units as possible in each case. We found that in most cases the classification of the testing example is not stable; it varies with different random seeds that initialize the networks. Table 2 summarises the experimental results. For ANNs, various classifications obtained by 20 different random seeds are listed with the first ones occurring most frequently. It seems that not only do neural networks not have a consistent default strategy, but also that it is neither the majority default as in ID3 nor the passthrough default as in SPA. This may explain why connectionist models cannot generalize unseen regular verbs well even when the training set contains only regular verbs (see Section 4.4). The networks have difficulty (or are underconstrained) in generalizing the identity mapping that copies the attributes of the verb stems into the past tenses.

| The classification for the testing example | | |
|---|---|---|
| | Data set 1 | Data set 2 |
| ID3 | $a$ | $c$ |
| SPA | $d$ | $c$ |
| ANN, dense coding | $b, c$ | $b$ |
| ANN, one-per-class | $b, c, a$ | $c, b$ |
| ANN, real numbers | $c, d$ | $d, c$ |

Table 2: Default strategies of ID3, SPA and ANN on two data sets.

## 4. Head-to-head Comparisons between Symbolic and ANN Models

In this section, we perform a series of extensive head-to-head comparisons using several different representations and encoding methods, and demonstrate that the SPA generalizes the past tense of unseen verbs better than ANN models do by a wide margin.

### 4.1 Format of the data

Our verb set came from MacWhinney's original list of verbs. The set contains about 1400 stem/past tense pairs. Learning is based upon the phonological UNIBET representation (MacWhinney, 1990), in which different phonemes are represented by different alphabetic/numerical letters. There is a total of 36 phonemes. The source file is transferred into the standard format of pairs of input and output patterns. For example, the verbs in Table 3 are represented as pairs of input and output patterns (verb stem => past tense):

```
6,b,&,n,d,6,n   =>   6,b,&,n,d,6,n,d
I,k,s,E,l,6,r,e,t  =>  I,k,s,E,l,6,r,e,t,I,d
```

```
6,r,3,z  => 6,r,o,z
b,I,k,6,m  =>  b,I,k,e,m
```

See Table 3 (The original verb set is available in Online Appendix 1). We keep only one form of the past tense among multiple past tenses (such as *hang-hanged* and *hang-hung*) in the data set. In addition, no homophones exist in the original data set. Consequently, there is no noise (contradictory data which have the same input pattern but different output patterns) in the training and testing examples. Note also that information as to whether the verb is regular or irregular is *not* provided in training/testing processes.

| base (stem) spelling form | UNIBET phonetic form | b=base d= past tense | 1 = irregular 0 = regular |
|---|---|---|---|
| abandon | 6b&nd6n | b | 0 |
| abandoned | 6b&nd6nd | d | 0 |
| benefit | bEn6fIt | b | 0 |
| benefited | bEn6fItId | d | 0 |
| arise | 6r3z | b | 0 |
| arose | 6roz | d | 1 |
| become | bIk6m | b | 0 |
| became | bIkem | d | 1 |
| ...... | | | |

Table 3: Source file from MacWhinney and Leinbach.

### 4.2 Experiment Setup

To guarantee unbiased and reliable comparison results, we use training and testing samples randomly drawn in several independent runs. Both SPA and ANN are provided with the *same* sets of training/testing examples for each run. This allows us to achieve a reliable estimate of the inductive generalization capabilities of each model on this task.

The neural network program we used is a package called Xerion, which was developed at the University of Toronto. It has several more sophisticated search mechanisms than the standard steepest gradient descent method with momentum. We found that training with the conjugate-gradient method is much faster than with the standard backpropagation algorithm. Using the conjugate-gradient method also avoids the need to search for proper settings of parameters such as the learning rate. However, we do need to determine the proper number of hidden units. In the experiments with ANNs, we first tried various numbers of hidden units and chose the one that produced the best predictive accuracy in a trial run, and then use the network with that number of hidden units in the actual runs. The SPA, on the other hand, has no parameters to adjust.

One major difference in implementation between ANNs and SPA is that SPA can take (symbolic) phoneme letters directly while ANNs normally encode each phoneme letter to binary bits. (Of course, SPA also can apply to the binary representation). We studied various binary encoding methods and compared results with SPA using symbolic letter representation. Since outputs of neural networks are real numbers, we need to decode the network outputs back to phoneme letters. We used the standard method of decoding: the phoneme letter that has the minimal real-number Hamming distance (smallest angle) with the network outputs was chosen. To see how binary encoding affects the generalization, the SPA was also trained with the binary representation. Since the SPA's outputs are binary, the decoding process may tie with several phoneme letters. In this case, one of them is chosen randomly. This reflects the probability of the correct decoding at the level of phoneme letters. When all of the phoneme letters are decoded, if one or more letters are incorrect, the whole pattern is counted as incorrect at the word level.

### 4.3 Templated, Distributed Representation

This set of experiments was conducted using the distributed representation suggested by MacWhinney and Leinbach (1991). According to MacWhinney and Leinbach, the output is a left-justified template in the format of CCCVVCCCVVCCCVVCCC, where C stands for consonant and V for vowel space holders. The input has two components: a left-justified template in the same format as the input, and a right-justified template in the format of VVCCC. For example, the verb *bet*, represented in UNIBET coding as *bEt*, is shown in the template format as follows (_ is the blank phoneme):

```
INPUT
bEt          b__E_t____________      _E__t
template:    CCCVVCCCVVCCCVVCCC      VVCCC
             (left-justified)        (right-justified)
OUTPUT
bEt          b__E_t____________
template:    CCCVVCCCVVCCCVVCCC
             (left-justified)
```

A specific distributed representation — a set of (binary) phonetic features — is used to encode all phoneme letters for the connectionist networks. Each vowel (V in the above templates) is encoded by 8 phonetic features (front, centre, back, high, low, middle, round, and diphthong) and each consonant (C in the above templates) by 10 phonetic features (voiced, labial, dental, palatal, velar, nasal, liquid, trill, fricative and interdental). Note that because the two feature sets of vowels and consonants are not identical, templates are needed in order to decode the right type of the phoneme letters from the outputs of the network.

In our experimental comparison, we decided not to use the right-justified template (VVCCC) since this information is redundant. Therefore, we used only the left-justified template (CCCVVCCCVVCCCVVCCC) in both input and output. (The whole verb set in the templated phoneme representation is available in Online Appendix 1. It contains 1320 pairs of verb stems and past tenses that fit the template). To ease implementation, we added two extra features that always were assigned to 0 in the vowel phonetic feature set. Therefore, both vowels and consonants were encoded by 10 binary bits. The ANN thus had $18 \times 10 = 180$ input bits and 180 output bits, and we found that one layer of 200 hidden units (same as MacWhinney (1993) model) reached the highest predictive accuracy in a trial run. See Figure 2 for the network architecture used.

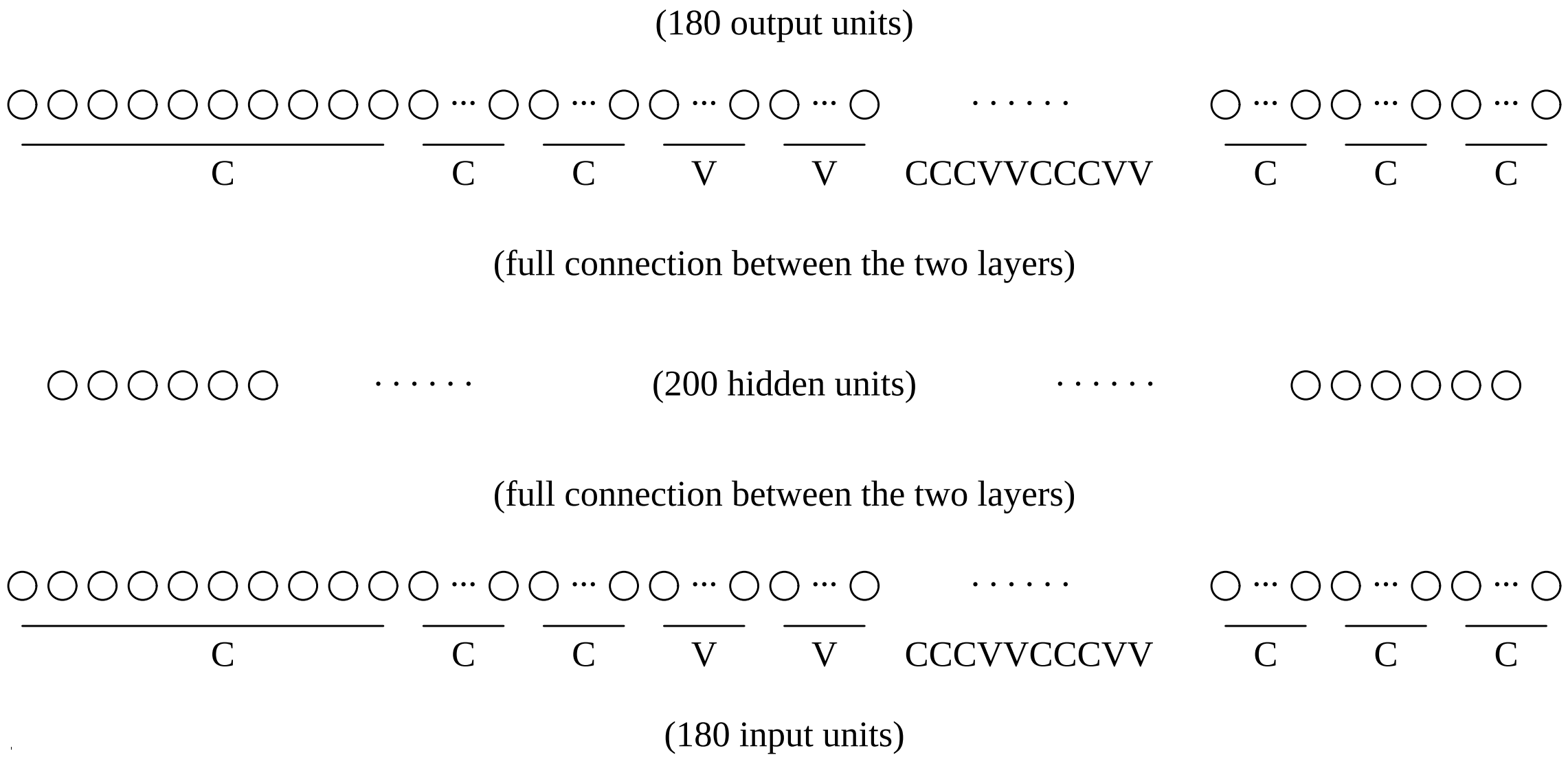

Figure 2: The architecture of the network used in the experiment.

The SPA was trained and tested on the same data sets but with phoneme letters directly; that is, 18 decision trees were built for each of the phoneme letters in the output templates. To see how phonetic feature encoding affects the generalization, we also trained the SPA with the the same distributed representation — binary bit patterns of 180 input bits and 180 output bits — exactly the same as those in the ANN simulation. In addition, to see how the "symbolic" encoding works in ANN, we also train another neural network (with 120 hidden units) with the "one-per-class" encoding. That is, each phoneme letter (total of 37; 36 phoneme letters plus one for blank) is encoded by 37 bits, one for each phoneme letter. We used 500 verb pairs (including both regular and irregular verbs) in the training and testing sets. Sampling was done randomly without replacement, and training and testing sets were disjoint. Three runs of SPA and ANN were conducted, and both SPA and ANN were trained and tested on the same data set in each run. Training reached 100% accuracy for SPA and around 99% for ANN.

Testing accuracy on novel verbs produced some interesting results. The ANN model and the SPA using the distributed representation have very similar accuracy, with ANN slightly better. This may well be caused by the binary outputs of SPA that suppress the fine differences in prediction. On the other hand, the SPA using phoneme letters directly produces much higher accuracy on testing. The SPA outperforms neural networks (with either distributed or one-per-class representations) by 20 percentage points! The testing results of ANN and SPA can be found in Table 4. Our findings clearly indicate that the SPA using symbolic representation leads to much better generalization than ANN models.

### 4.4 Learning Regular Verbs

Predicting the past tense of an unseen verb, which can be either regular or irregular, is not an easy task. Irregular verbs are not learned by rote as traditionally thought since children and adults occasionally extend irregular inflection to irregular-sounding regular verbs or pseudo verbs (such as *cleef — cleft*) (Prasada & Pinker, 1993). The more similar the novel verb is to the cluster of irregular verbs with similar phonological patterns, the more likely the prediction of an irregular past-tense form. Pinker (1991) and Prasada and Pinker (1993) argue that regular past tenses are governed by rules, while irregulars may be generated by the associated memory which has this graded effect of irregular past-tense generalization. It is would be interesting, therefore, to compare SPA and ANN on the past-tense generalization of regular verbs only. Because both SPA and ANN use the same, position specific, representation, learning regular past tenses would require learning different suffixes[2] at *different* positions, and to learn the identity mapping that copies the verb stem to the past tenses for verbs of different lengths.

| Distributed representation | | | | | | Symbolic representation | | | | | |
|---|---|---|---|---|---|---|---|---|---|---|---|
| ANN: % Correct | | | SPA: % Correct | | | ANN: % Correct | | | SPA: % Correct | | |
| Reg | Irrg | Comb | Reg | Irrg | Comb | Reg | Irrg | Comb | Reg | Irrg | Comb |
| 65.3 | 14.6 | 60.4 | 62.2 | 18.8 | 58.0 | 63.3 | 18.8 | 59.2 | 83.0 | 29.2 | 77.8 |
| 59.7 | 8.6 | 53.8 | 57.9 | 8.2 | 52.2 | 58.8 | 10.3 | 53.2 | 83.3 | 22.4 | 76.2 |
| 60.0 | 16.0 | 55.6 | 58.0 | 8.0 | 53.0 | 58.7 | 16.0 | 54.4 | 80.9 | 20.0 | 74.8 |
| 61.7 | 13.1 | 56.6 | 59.4 | 11.7 | 54.4 | 60.3 | 15.0 | 55.6 | 82.4 | 23.9 | 76.3 |

Table 4: Comparisons of testing accuracy of SPA and ANN with distributed and symbolic representations.

We used the same templated representation as in the previous section, but both training and testing sets contained only regular verbs. Again samples were drawn randomly without replacement. To maximize the size of the testing sets, testing sets simply consisted of all regular verbs that were not sampled in the training sets. The same training and testing sets were used for each of the following methods compared. To see the effect of the adaptive default strategy (as discussed in Section 3.2) on generalization, the SPA with the majority default only and with the adaptive default were both tested. The ANN models were similar to those used in the previous section (except with 160 one-layer hidden units, which turned out to have the best predictive accuracy in a test run). The passthrough default strategy can be imposed on neural networks by adding a set of copy connections that connect directly from the input units to the twin output units. MacWhinney and Leinbach (1991) used such copy connections in their simulation. We therefore tested the networks with the copy connection to see if generalization would be improved as well.

The results on the predictive accuracy of the SPA and ANNs on one run with with randomly sampled training and testing sets are summarized in Table 5. As we can see, the SPA with the adaptive default strategy, which combines the majority and passthrough default, outperforms the SPA with only the majority default strategy used in ID3. The

2. In phonological form there are three different suffixes for regular verbs. When the verb stem ends with *t* or *d* (UNIBET phonetic representations), then the suffix is *Id*. For example, *extend — extended* (in spelling form). When the verb stem ends with a unvoiced consonant, the suffix is *t*. For example, *talk — talked*. When the verb stem ends with a voiced consonant or vowel, the suffix is *d*. For example, *solve — solved*.

ANNs with copy connections do generalize better than the ones without. However, even ANN models with copy connections have a lower predictive accuracy than the SPA (majority). In addition, the differences in the predictive accuracy are larger with smaller sets of training examples. Smaller training sets make the difference in testing accuracy more evident. When the training set contains 1000 patterns (out of 1184), the testing accuracy becomes very similar, and would approach asymptotically to 100% with larger training sets. Upon examination, most of the errors made in ANN models occur in the identity mapping (i.e., strange phoneme change and drop); the verb stems cannot be preserved in the past tense if the phonemes are not previously seen in the training examples. This contradicts the findings of Prasada and Pinker (1993), which show that native English speakers generate regular suffix-adding past tenses equally well with unfamiliar-sounding verb stems (as long as these verb stems do not sound close to irregular verbs). This also indicates that the bias of the ANN learning algorithms is not suitable to this type of task. See further discussion in Section 5.

| Training size | Percent correct on testing | | | |
|---|---|---|---|---|
| | SPA (adaptive) | SPA (majority) | ANN (copy con.) | ANN (normal) |
| 50 | 55.4 | 30.0 | 14.6 | 7.3 |
| 100 | 72.9 | 58.6 | 34.6 | 24.9 |
| 300 | 87.0 | 83.7 | 59.8 | 58.2 |
| 500 | 92.5 | 89.0 | 82.6 | 67.9 |
| 1000 | 93.5 | 92.4 | 92.0 | 87.3 |

Table 5: Predictive accuracy on learning the past tense of regular verbs

### 4.5 Error Correcting Codes

Dietterich and Bakiri (1991) reported an increase in the predictive accuracy when error-correcting codes of large Hamming distances are used to encode values of the attributes. This is because codes with larger Hamming distance ($d$) allow for correcting fewer than $d/2$ bits of errors. Thus, learning programs are allowed to make some mistakes at the bit level without their outputs being misinterpreted at the word level.

We wanted to find if performances of the SPA and ANNs are improved with the error-correcting codes encoding all of the 36 phonemes. We chose error-correcting codes ranging from ones with small Hamming distance to ones with very large Hamming distance (using the BHC codes, see Dietterich and Bakiri (1991)). Because the number of attributes for each phoneme is too large, the data representation was changed slightly for this experiment. Instead of 18 phoneme holders with templates, 8 consecutive, left-to-right phoneme holders were used. Verbs with stems or past tenses of more than 8 phonemes were removed from the training/testing sets. (The whole verb set in the this representation is available in Online Appendix 1. It contains 1225 pairs of verb stems and past tenses whose lengths are shorter than 8). Both SPA and ANN take exactly the same training/testing sets, each contains 500 pairs of verb stems and past tenses, with the error-correcting codes encoding each phoneme letter. Still, training networks with 92-bit or longer error-correcting codes takes too long to run (there are $8 \times 92 = 736$ input attributes and 736 output attributes). Therefore, only two runs with 23- and 46-bit codes were conducted. Consistent with Dietterich and Bakiri (1991)'s findings, we found that the testing accuracy generally increases when the Hamming distance increases. However, we also observed that the testing accuracy decreases very slightly when the codes become too long. The accuracy using 46-bit codes (with Hamming distance of 20) reaches the maximum value (77.2%), which is quite close to the accuracy (78.3%) of SPA using the direct phoneme letter representation. It seems there is a trade-off between tolerance of errors with large Hamming distance and difficulty in learning with longer codes. In addition, we found the testing accuracy of ANNs to be lower than the one of SPA for both 23 bit- and 46-bit error-correcting codes. The results are summarized in Table 6.

| **ANN** | Hamming Distance | Correct at bit level | Correct at word level |
|---|---|---|---|
| 23-bit codes | 10 | 93.5% | 65.6% |
| 46-bit codes | 20 | 94.1% | 67.4% |
| **SPA** | Hamming Distance | Correct at bit level | Correct at word level |
| 23-bit codes | 10 | 96.3% | 72.4% |
| 46-bit codes | 20 | 96.3% | 77.2% |
| 92-bit codes | 40 | 96.1% | 75.6% |
| 127-bit codes | 54 | 96.1% | 75.4% |

Table 6: Comparisons of the testing accuracy of SPA and ANNs with error-correcting codes

Our results in this and the previous two subsections undermine the advantages of the distributed representation of ANNs, a unique feature advocated by connectionists. We have demonstrated that, in this task, the distributed representation actually does not allow for adequate generalization. Both SPA using direct symbolic phoneme letters and SPA with error-correcting codes outperform ANNs with distributed representation by a wide margin. However, neither phoneme symbols nor bits in the error-correcting codes encode, *implicitly* or *explicitly*, any micro-features as in the distributed representation. It may be that the distributed representation used was not optimally designed. Nevertheless, straightforward symbolic format requires little representation engineering compared with the distributed representation in ANNs.

### 4.6 Right-justified, Isolated Suffix Representation

MacWhinney and Leinbach (1991) did not report important results of the predictive accuracy of their model on unseen regular verbs. In his reply (MacWhinney, 1993) to our paper (Ling & Marinov, 1993), MacWhinney re-implemented the ANN model. In his new implementation, 1,200 verb stem and past-tense pairs were in the training set, among which 1081 were regular and 119 were irregular. Training took 4,200 epochs, and reached 100% correct on regulars and 80% on irregulars. The testing set consisted of 87 regulars and 15 irregulars. The percent correct on testing at epoch 4,200 was 91% for regulars and 27% for irregulars, with a combined 80.0% on the testing set. MacWhinney claimed that the raw generalization power of ANN model is very close to that of our SPA. He believes that this should be the case simply because both systems were trained on the same data set.

We realize (via private communication) that a new representation used in MacWhinney's recent implementation plays a critical role in the improved performance. In MacWhinney's new representation, the input (for verb stems) is coded by the *right*-justified template CCCVVCCCVVCCCVVCCC. The output contains two parts: a right-justified template that is the same as the one in the input, and a coda in the form of VVCCC. The right-justified template in the output is used to represent the past tense *without* including the suffix for the regular verbs. The suffix of the regular past tense always stays in the coda, which is *isolated* from the main, right-justified templates. For the irregular past tense, the coda is left empty. For example, the input and output templated patterns for the past tense of verbs in Table 3 are represented as:

```
INPUT                 OUTPUT
(right-justified)     (right-justified)   (suffix only)
CCCVVCCCVVCCCVVCCC    CCCVVCCCVVCCCVVCCC  VVCCC
___6_b__&_nd_6_n__    ___6_b__&_nd_6_n__  __d__ (for abandon-abandoned)
b__E_n__6_f__I_t__    b__E_n__6_f__I_t__  I_d__ (for benefit-benefited)
________6_r__3_z__    ________6_r__o_z__  _____ (for arise-arose)
_____b__I_k__6_m__    _____b__I_k__e_m__  _____ (for become-became)
```

Such data representation clearly facilitates learning. For the regular verbs, the output patterns are always identical to the input patterns. In addition, the verb-ending phoneme letters always appear at a few fixed positions (i.e., the right most VVCCC section in the input template) due to the right-justified, templated representation. Furthermore, the suffix always occupies the coda, isolated from the right-justified templates.

We performed a series of experiments to see how much improvement we could accomplish using the new representation over MacWhinney's recent ANN model and over the left-justified representation discussed in Section 4.3. Our SPA (with an averaged predictive accuracy of 89.0%) outperforms MacWhinney's recent ANN implementation (with the predictive accuracy of 80.0%) by a wide margin. In addition, the predictive accuracy is also improved from an average of 76.3% from the left-justified representation to 82.8% of the right-justified, isolated suffix one. See results in Table 7.

## 5. General Discussion and Conclusions

Two factors contribute to the generalization ability of a learning program. The first is the data representation, and the other is the bias of the learning program. Arriving at the right, optimal, representation is a difficult task. As argued by Prasada and Pinker (1993), regular verbs should be represented in a coarse grain in terms of the verb stem and suffixes; while irregular verbs in a finer grain in terms of phonological properties. Admittedly, SPA works uniformly at the level of phoneme letters, as ANNs do. However, because SPA produces simple production rules that use these phoneme letters directly, those rules can be further generalized to first-order rules with new representations such as stems and the voiced consonants which can be used across the board in other such rule-learning modules (Ling & Marinov, 1993). This is one of the major advantages over ANN models.

| | Predictive accuracy with right-justified, isolated suffix representation | | |
|---|---|---|---|
| | SPA | | MacWhinney's ANN model |
| | training/testing 500/500 | training/testing 1200/102 | training/testing 1200/102 |
| Run 1 | 81.3 | 89.2 | |
| Run 2 | 84.1 | 90.4 | |
| Run 3 | 83.1 | 87.4 | |
| Average | 82.8 | 89.0 | 80.0 (one run) |

Table 7: Comparisons of testing accuracy of SPA and ANN (with right-justified, isolated suffix representation)

It seems quite conceivable that children acquire these high-level concepts such as stems and voiced consonants through learning noun plurals, verb past tense, verb third-person singular, comparative adjectives, and so on. With a large weight matrix as the result of learning, it is hard to see how this knowledge can be further generalized in ANN models and shared in other modules.

Even with exactly the same data representation, there exist some learning tasks that symbolic methods such as the SPA generalize categorically better than ANNs. The converse also is true. This fact reflects the different inductive biases of the different learning algorithms. The Occam's Razor Principle — preferring the simplest hypothesis over more complex ones — creates a preference bias, a preference of choosing certain hypotheses over others in the hypothesis space. However, different learning algorithms choose different hypotheses because they use different measurements for simplicity. For example, among all possible decision trees that fit the training examples, ID3 and SPA induce simple decision trees instead of complicated ones. Simple decision trees can be converted to small sets of production rules. How well a learning algorithm generalizes depends upon the degree to which the underlying regularities of the target concept fit its bias. In other words, if the underlying regularities can be represented *compactly* in the format of hypotheses produced by the learning algorithm, the data can be generalized well, even with a small set of training examples. Otherwise, if the underlying regularities only have a large hypothesis, but the algorithm is looking for compact ones (as per the Occam's Razor Principle), the hypotheses inferred will not be accurate. A learning algorithm that searches for hypotheses larger than necessary (i.e., that does not use the Occam's Razor Principle) is normally "underconstrained"; it does not know, based on the training examples only, which of the many competitive hypotheses of the large size should be inferred.

We also can describe the bias of a learning algorithm by looking at how training examples of different classes are separated in the $n$-dimensional hyperspace where $n$ is the number of attributes. A decision node in a decision tree forms a hyperplane as described by a linear function such as $X = a$. Not only are these hyperplanes *perpendicular* to the axis, they are also *partial-space* hyperplanes that extend only within the subregion formed by the hyperplanes of their parents' nodes. Likewise, hidden units with a threshold function in ANNs can be viewed as forming hyperplanes in the hyperspace. However, unlike the ones in the decision trees, they need not be perpendicular to any axis, and they are *full-space* hyperplanes that extend through the whole space. If ID3 is applied to the concepts that fit ANN's bias, especially if their hyperplanes are not perpendicular to any axis, then many *zigzag* hyperplanes that are perpendicular to axes would be needed to separate different classes of the examples. Hence, a large decision tree would be needed, but this does not fit ID3's bias. Similarly, if ANN learning algorithms are applied to the concepts that fit ID3's bias, especially if their hyperplanes form many separated, partial-space regions, then many hidden units may be needed for these regions.

Another major difference between ANNs and ID3 is that ANNs have a larger variation and a weaker bias (cf. (Geman, Bienenstock, & Doursat, 1992)) than ID3. Many more Boolean functions (e.g., linearly separable functions) can fit a small network (e.g., one with no hidden units) than they can a small decision tree. This is sometimes attributed to the claimed versatility and flexibility of ANNs; they can learn (but not necessarily predict reliably well) many functions, while symbolic methods are brittle. However, it is my belief that we humans are versatile, not because we have a learning algorithm with a large variation, but rather because we have *a set of strong-biased* learning algorithms, and we can somehow search in the bias space and add new members into the set for the new learning tasks. Symbolic learning algorithms have clear semantic components and explicit representation, and thus we can more easily construct strong-based algorithms motivated from various specific learning tasks. The adaptive default strategy in the SPA is such an example. On the other hand, we still largely do not know *how* to effectively strengthen the bias of ANNs for many specific tasks (such as the identity mapping, $k$-term DNF, etc.). Some techniques (such as adding copy connections and weight decaying) exist, but their exact effects on biasing towards classes of functions are not clear.

From our analyses (Ling & Marinov, 1993), the underlying regularities governing the inflection of the past tense of English verbs do form a small set of production rules with phoneme letters. This is especially so for regular verbs; all the rules are either identity rules or the suffix-adding rules. For example, decision trees can be converted into a set of *precedence-ordered* production rules with more complicated rules (rules with more conditions) listed first. As an example, using consecutive, left-to-right phonetic representation, a typical suffix-adding rule for verb stems with 4 phoneme letters (such as *talk* — *talked*) is:

If $\iota_4$ = k and $\iota_5$ = _, then $\omega_5$ = t

That is, if the fourth input phoneme is *k* and the fifth is blank (i.e., if we are at a verb ending) then the fifth output phoneme is *t*. On the other hand, the identity-mapping rules have only one condition. A typical identity rule looks like:

If $\iota_3$ = l, then $\omega_3$ = l

In fact, the passthrough default strategy allows all of the identity-mapping rules to be represented in a simple first-order format:

If $\iota_3 = \mathbf{X}$, then $\omega_3 = \mathbf{X}$

where **X** can be any phoneme. Clearly, the knowledge of forming regular past tenses can thus be expressed in simple, conjunctive rules which fit the bias of the SPA (ID3), and therefore, the SPA has a much better generalization ability than the ANN models.

To conclude, we have demonstrated, via extensive head-to-head comparisons, that the SPA has a more realistic and better generalization capacity than ANNs on learning the past tense of English verbs. We have argued that symbolic decision-tree/production-rule learning algorithms should outperform ANNs. This is because, first, the domain seems to be governed by a compact set of rules, and thus fits the bias of our symbolic learning algorithm; second, the SPA directly manipulates on a representation better than ANNs do (i.e., the symbolic phoneme letters vs. the distributed representation); and third, the SPA is able to derive high-level concepts used throughout English morphology. Our results support the view that many such high-level, rule-governed cognitive tasks should be better modeled by symbolic, rather than connectionist, systems.

## Acknowledgements

I gratefully thank Steve Pinker for his constant encouragement, and Marin Marinov, Steve Cherwenka and Huaqing Zeng for discussions and for help in implementing the SPA. I thank Brian MacWhinney for providing the verb data used in his simulation. Discussions with Tom Dietterich, Dave Touretzky and Brian MacWhinney, as well as comments from reviewers, have been very helpful. The research is conducted with support from the NSERC Research Grant and computing facilities from our Department.

## References

Cottrell, G., & Plunkett, K. (1991). Using a recurrent net to learn the past tense. In *Proceedings of the Cognitive Science Society Conference*.

Daugherty, K., & Seidenberg, M. (1993). Beyond rules and exceptions: A connectionist modeling approach to inflectional morphology. In Lima, S. (Ed.), *The Reality of Linguistic Rules*. John Benjamins.

Dietterich, T., & Bakiri, G. (1991). Error-correcting output codes: A general method for improving multiclass inductive learning programs. In *AAAI-91 (Proceedings of Ninth National Conference on Artificial Intelligence)*.

Dietterich, T., Hild, H., & Bakiri, G. (1990). A comparative study of ID3 and backpropagation for English text-to-speech mapping. In *Proceedings of the 7th International Conference on Machine Learning*. Morgan Kaufmann.

Feng, C., King, R., Sutherland, A., & Henery, R. (1992). Comparison of symbolic, statistical and neural network classifiers. Manuscript. Department of Computer Science, University of Ottawa.

Fodor, J., & Pylyshyn, Z. (1988). Connectionism and cognitive architecture: A critical analysis. In Pinker, S., & Mehler, J. (Eds.), *Connections and Symbols*, pp. 3 – 71. Cambridge, MA: MIT Press.

Geman, S., Bienenstock, E., & Doursat, R. (1992). Neural networks and the bias/variance dilemma. *Neural Computation*, *4*, 1 – 58.

Lachter, J., & Bever, T. (1988). The relation between linguistic structure and associative theories of language learning – a constructive critique of some connectionist learning models. In Pinker, S., & Mehler, J. (Eds.), *Connections and Symbols*, pp. 195 – 247. Cambridge, MA: MIT Press.

Ling, X., Cherwenka, S., & Marinov, M. (1993). A symbolic model for learning the past tenses of English verbs. In *Proceedings of IJCAI-93 (Thirteenth International Conference on Artificial Intelligence)*, pp. 1143–1149. Morgan Kaufmann Publishers.

Ling, X., & Marinov, M. (1993). Answering the connectionist challenge: a symbolic model of learning the past tense of English verbs. *Cognition*, *49*(3), 235–290.

MacWhinney, B. (1990). *The CHILDES Project: Tools for Analyzing Talk.* Hillsdale, NJ: Erlbaum.

MacWhinney, B. (1993). Connections and symbols: closing the gap. *Cognition*, *49*(3), 291–296.

MacWhinney, B., & Leinbach, J. (1991). Implementations are not conceptualizations: Revising the verb model. *Cognition*, *40*, 121 – 157.

Pinker, S. (1991). Rules of language. *Science*, *253*, 530 – 535.

Pinker, S., & Prince, A. (1988). On language and connectionism: Analysis of a parallel distributed processing model of language acquisition. In Pinker, S., & Mehler, J. (Eds.), *Connections and Symbols*, pp. 73 – 193. Cambridge, MA: MIT Press.

Plunkett, K., & Marchman, V. (1991). U-shaped learning and frequency effects in a multilayered perceptron: Implications for child language acquisition. *Cognition*, *38*, 43 – 102.

Prasada, S., & Pinker, S. (1993). Generalization of regular and irregular morphological patterns. *Language and Cognitive Processes*, *8*(1), 1 – 56.

Quinlan, J. (1986). Induction of decision trees. *Machine Learning*, *1*(1), 81 – 106.

Quinlan, J. (1993). *C4.5 Programs for Machine Learning.* Morgan Kaufmann: San Mateo, CA.

Ripley, B. (1992). Statistical aspects of neural networks. Invited lectures for SemStat (Seminaire Europeen de Statistique, Sandbjerg, Denmark, 25-30 April 1992).

Rumelhart, D., & McClelland, J. (1986). On learning the past tenses of English verbs. In Rumelhart, D., McClelland, J., & the PDP Research Group (Eds.), *Parallel Distributed Processing Vol 2*, pp. 216 – 271. Cambridge, MA: MIT Press.

Shavlik, J., Mooney, R., & Towell, G. (1991). Symbolic and neural learning algorithms: An experimental comparison. *Machine Learning*, *6*(2), 111 – 144.

Weiss, S., & Kulikowski, C. (1991). *Computer Systems that Learn: classification and prediction methods from statistics, neural networks, machine learning, and expert systems.* Morgan Kaufmann, San Mateo, CA.